\documentclass[11pt, a4paper]{article}

\usepackage[margin=1in, top=1.2in, bottom=1in]{geometry}

\usepackage{iftex}
\ifPDFTeX
  \usepackage[T1]{fontenc}
  \usepackage[utf8]{inputenc}
  \usepackage{lmodern}
\else
  \usepackage{fontspec}
\fi

\usepackage{graphicx}
\usepackage[table]{xcolor}
\usepackage{booktabs}
\usepackage{longtable}
\usepackage{tabularx}
\usepackage{float}
\usepackage{caption}
\usepackage{subcaption}
\usepackage{enumitem}
\usepackage{multicol}
\usepackage{titlesec}
\usepackage{listings}
\usepackage{textcomp}
\usepackage{hyperref}
\usepackage{fancyhdr}
\usepackage[export]{adjustbox}
\usepackage{amsmath}
\usepackage{amssymb}
\usepackage{lastpage}

\usepackage[authoryear,round]{natbib}
\usepackage{pgfplots}
\pgfplotsset{compat=1.18}
\definecolor{portexprimary}{HTML}{4A90D9}
\definecolor{portexsecondary}{HTML}{7B68EE}
\definecolor{portexsuccess}{HTML}{27AE60}
\definecolor{portexfail}{HTML}{E74C3C}
\definecolor{portexwarn}{HTML}{F39C12}
\definecolor{portexaccent}{HTML}{D97706}
\definecolor{portexgray}{HTML}{7F8C8D}
\definecolor{portexlight}{HTML}{ECF0F1}
\definecolor{portexcached}{HTML}{95A5A6}
\definecolor{portexinput}{HTML}{3498DB}
\definecolor{portexoutput}{HTML}{E67E22}
\definecolor{portexlink}{HTML}{1E3A8A}

\hypersetup{
    colorlinks=true,
    linkcolor=portexlink,
    citecolor=portexlink,
    urlcolor=portexlink,
}

\fancypagestyle{firstpage}{
    \fancyhf{}
    \fancyfoot[C]{%
        \begin{minipage}{0.96\textwidth}
        \centering
        \small\textcolor{portexgray}{Page \thepage\ of \pageref{LastPage}}
        \end{minipage}
    }
    
}

\titleformat{\section}{\large\bfseries}{\thesection}{0em}{}
\titleformat{\subsection}{\normalsize\bfseries}{\thesubsection}{0em}{}
\titlespacing*{\section}{0pt}{0.7ex plus 0.2ex minus 0.1ex}{0.35ex plus 0.1ex}
\titlespacing*{\subsection}{0pt}{0.55ex plus 0.2ex minus 0.1ex}{0.25ex plus 0.1ex}

\newsavebox{\taskbox}

\begin{document}

\thispagestyle{firstpage}
\begin{center}
\vspace*{-2.5cm}
\includegraphics[width=1.2cm]{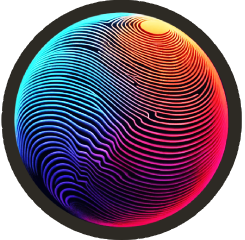}
\vspace{0.08cm}

{\LARGE\bfseries COMPOSITE-STEM}\\[0.2em]
\vspace{0.02em}
\small 70 expert-curated agentic tasks across Physics, Biology, Chemistry, and Math.\\[0.06em]
\begin{tikzpicture}
  \shade[left color=portexaccent, right color=portexprimary!40]
    (0,0) rectangle (\textwidth,0.03);
  \shade[left color=portexaccent!85, right color=portexprimary!30]
    (0,-0.045) rectangle (\textwidth,-0.075);
\end{tikzpicture}\\[0.06em]
\end{center}

\vspace{0.03em}
\begin{center}
{\large\bfseries Portex Organizing Team}\\[0.10em]
\begin{tabularx}{0.95\textwidth}{>{\centering\arraybackslash}X >{\centering\arraybackslash}X >{\centering\arraybackslash}X}
\textbf{Kyle Waters} & \textbf{Lucas Nuzzi} & \textbf{Tadhg Looram} \\
PortexAI & PortexAI & PortexAI \\
\texttt{kyle@portexai.com} & \texttt{lucas@portexai.com} & \texttt{tadhg@portexai.com}
\end{tabularx}\\[1.5em]
{\large April 2026}
\end{center}

\vspace{-0.40em}
\begin{center}
\textbf{\large Dataset Contributors}\\[0.08em]
\textbf{Alessandro Tomasiello} (University of Milano-Bicocca), \textbf{Ariel Ghislain Kemogne Kamdoum} (University of Calgary), \textbf{Bikun Li} (University of Chicago), \textbf{Damien Sileo} (Inria), \textbf{Egor Kretov} (Fraunhofer Institute for Individualized Medical Technology IMTE), \textbf{Francesco Fournier-Facio} (University of Cambridge), \textbf{Georgios Soloupis} (Independent), \textbf{Haile Kassahun} (McGill University), \textbf{Hew Wolff} (Independent), \textbf{Jiaqi Cai} (Massachusetts Institute of Technology), \textbf{Lianghui Li} (École Polytechnique Fédérale de Lausanne), \textbf{Marc Roth} (Queen Mary University of London), \textbf{Mohinder Naiya} (Dot Ingredients), \textbf{Naixu Guo} (National University of Singapore), \textbf{Qicheng Tang} (Georgia Institute of Technology), \textbf{Richard Wheeler} (University of Edinburgh), \textbf{Samuele Sala} (Murdoch University), \textbf{Serguei Popov} (University of Porto), \textbf{Steven Dillmann} (Stanford University), \textbf{Yuqi Li} (Stony Brook University)
\end{center}
\vspace{-0.32em}
\begin{center}
\begin{minipage}{0.90\textwidth}
\footnotesize\textbf{Abstract.} AI agents hold growing promise for accelerating scientific discovery; yet, a lack of frontier evaluations hinders adoption into real workflows. Expert-written benchmarks have proven effective at measuring AI reasoning, but most at this stage have become saturated and only measure performance on constrained outputs. To help address this gap, we introduce \textbf{COMPOSITE-STEM}, a benchmark of 70 expert-written tasks in physics, biology, chemistry, and mathematics, curated by doctoral-level researchers. Our benchmark combines exact-match grading and criterion-based rubrics with an LLM-as-a-jury grading protocol, allowing more flexible assessment of scientifically meaningful outputs. Using an adapted multimodal Terminus-2 agent harness within the Harbor agentic evaluation framework, we evaluate four frontier models. The top-performing model achieves 21\%, demonstrating that COMPOSITE-STEM captures capabilities beyond current agent reach. All tasks are open-sourced with contributor permission to support reproducibility and to promote additional research towards AI's acceleration of scientific progress in these domains.
\end{minipage}
\end{center}
\vspace{0.04em}

\raggedcolumns
\setlength{\columnsep}{0.6cm}
\setlength{\multicolsep}{2pt}
\setlength{\premulticols}{50pt}
\setlength{\postmulticols}{0pt}
\begin{multicols}{2}

\section{Introduction}

Scientific evaluations are central to advancing frontier AI for real scientific workflows. In this work, we introduce \textbf{COMPOSITE-STEM}, a cross-domain STEM task bundle compatible with \href{https://harborframework.com/}{Harbor} (TerminalBench-style agentic evaluation). This paper documents benchmark construction, task curation, and model performance.

We evaluate 4 models using a modified Terminus-2 agent harness adapted for multimodal support in Harbor. claude-opus-4.6 leads at 21.4\% (Pass@1).

The benchmark is designed to test more than isolated scientific reasoning by pairing expert-authored tasks with executable environments and flexible grading.

\section{Leaderboard Snapshot}
\begin{center}
\includegraphics[width=0.98\columnwidth]{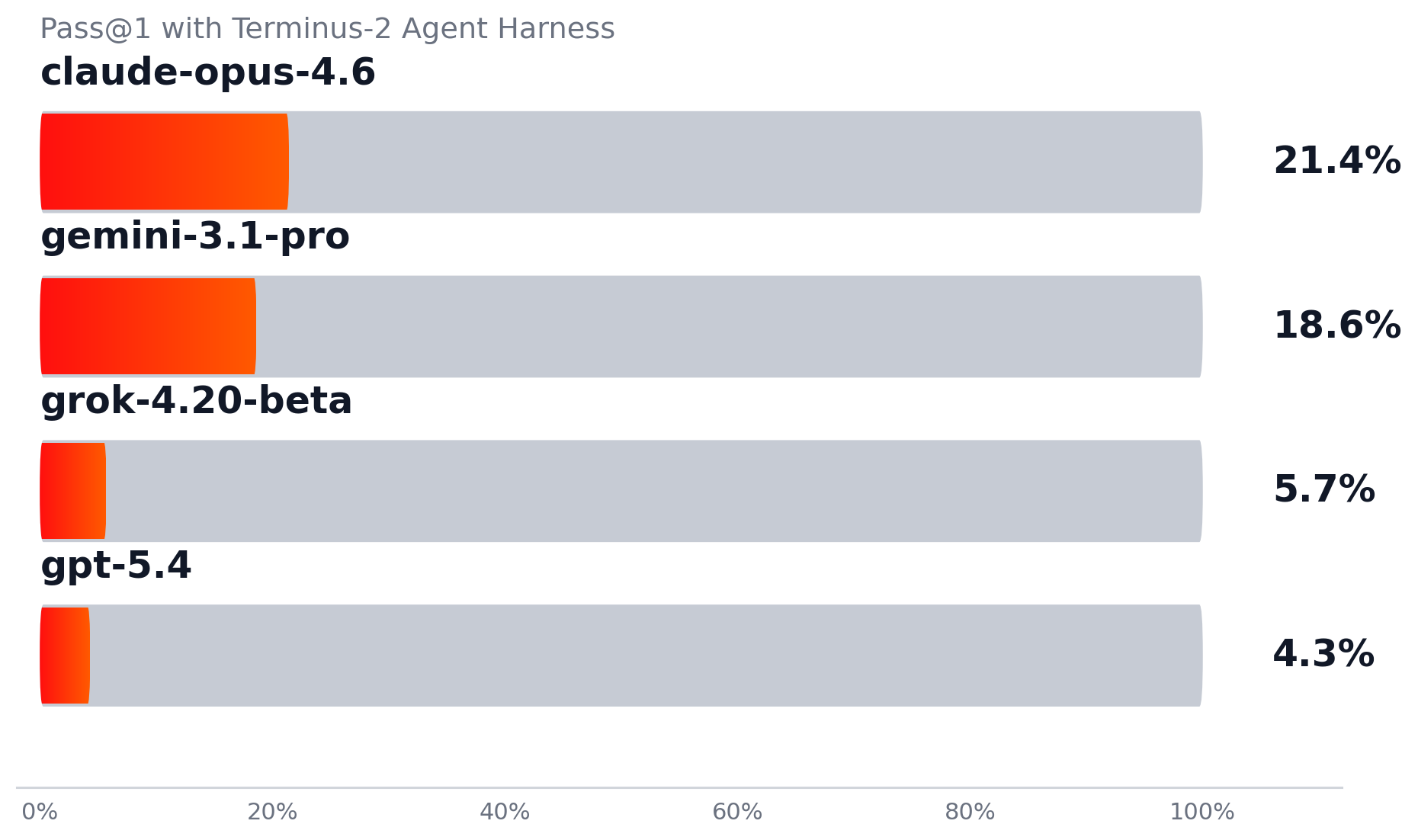}\\[-0.1em]
\footnotesize Figure 1: COMPOSITE-STEM leaderboard on 70 tasks (Pass@1).
\end{center}

\section{Related Work}

Benchmark design has been shifting from short, static reasoning tests toward harder, longer-horizon, and more realistic tasks that better reflect real-world work. This has been necessitated by the quick advancement in capabilities, coupled with the saturation and contamination of many expert-grade benchmarks. When GPQA \citet{gpqa} was introduced in 2023 as a benchmark of “Google-proof” multiple-choice science questions written by PhD-level experts, GPT‑4 scored 39\%, but only two years later, GPT‑5.2 scored 92\%.

A milestone in benchmark design was the release of Humanity's Last Exam or HLE, which established a multi-subject, academic-focused benchmark at scale. HLE covered over 2,500 questions from 500+ contributors sourced from top global research institutions, and revealed clear limits to SOTA model capabilities at the time of its release \citet{hle}. However, HLE remains primarily a static QA-style benchmark with exact-match and multiple-choice style evaluation. Moreover, despite efforts to prevent training on benchmark data, data contamination has become a real concern as HLE is a high-profile benchmark that has garnered attention and is cited in almost all model releases.

As agentic systems like Claude Code and Codex gained popularity, benchmark efforts increasingly moved into executable environments. Terminal-Bench Core and Terminal-Bench 2.0 were important steps in evaluating agents in terminals with reproducible containerized settings. The current 2.0 benchmark includes 89 curated tasks spanning software engineering, ML, security, and data workflows \citet{terminalbench2}. In parallel, GDPval expanded realism from an economic-work perspective with 1,320 professionally grounded tasks curated by domain experts, emphasizing high-value deliverables (e.g. excel spreadsheets, presentations, PDF documents) rather than only abstract reasoning \citet{gdpval}. Mercor contributes two distinct and relevant papers in this space. First, APEX-v1-extended evaluates economically valuable professional tasks (law, finance, consulting, and general medicine) with prompt-specific grading rubrics and LM-judge scoring, illustrating how rubric-based evaluation can support richer, less brittle assessment than exact-match-only benchmarks \citet{apex}. Second, APEX-Agents pushes further into fully agentic workflows with 480 tasks across realistic multi-application professional environments (called "worlds"), bringing rubric-based scoring into longer-horizon agent execution settings \citet{apexagents}.

Finally,  OpenAI's introduction of FrontierScience \citet{frontierscience} further advanced expert-sourced scientific benchmarking by pairing difficult, original science problems with a more structured evaluation framework for open-ended answers. The benchmark spans physics, chemistry, and biology, and is split into two tracks: an Olympiad track built from expert-written, verifiable short-answer problems, and a Research track composed of PhD-level research subproblems designed to reflect authentic scientific reasoning tasks. Most relevant here, FrontierScience’s Research track moves beyond exact-match grading by assigning each task a 10-point rubric made up of multiple independent, objectively assessable criteria, allowing evaluation of intermediate reasoning steps rather than only final-answer correctness. A response is treated as successful if it earns at least 7 out of 10 rubric points, enabling finer-grained analysis of partial progress and failure modes. To scale grading, FrontierScience uses a single LLM judge, GPT-5 with high reasoning effort, to score submissions against these expert-authored rubrics.

COMPOSITE-STEM sits at the intersection of these directions: expert-authored STEM tasks paired with reproducible Harbor-compatible agent execution. The goal is to preserve the academic rigor of difficult expert benchmarks while evaluating performance in realistic settings across physics, chemistry, biology, and math.

\section{Task Composition}

COMPOSITE-STEM contains 70 tasks: 20 in physics, 23 in chemistry, 20 in biology, and 7 in math. The tasks mix structured problem solving, multi-step reasoning, and domain-specific constraints, so strong performance requires both conceptual understanding and knowledge of execution (e.g. most appropriate Python packages).

Reference assets are also a meaningful part of the task mix: 18/70 tasks include files mounted into the agent sandbox (under \texttt{/app/refs}), and these are mostly images (17 image files, primarily PNG, plus 1 PDF).

All tasks are open sourced on \href{https://huggingface.co/datasets/portex/COMPOSITE-STEM/tree/main}{Hugging Face} while a \href{https://github.com/portex-ai/portex-composit-harbor-adapter}{Harbor adapter} is made available on GitHub.

\section{Expert Background}

Contributors for COMPOSITE-STEM were sourced from top research institutions and global universities and primarily included contributors to \href{https://agi.safe.ai/}{Humanity's Last Exam (HLE)} and adjacent frontier benchmarks. Across domains, the contributor group includes doctoral-level researchers, distinguished faculty members, postdoctoral scientists, and industry practitioners with publication records and prior benchmark design experience. Beyond sourcing, the Portex team worked closely with all contributors through detailed calls and review cycles to shape task design, clarify grading intent, and identify specification issues before release.

\section{Environment}

We use an adapted multimodal \href{https://harborframework.com/docs/agents/terminus-2}{Terminus-2} agent harness in the \href{https://www.tbench.ai/}{Terminal-Bench} and Harbor ecosystem, to minimize harness-specific variance while better supporting visual-based tasks such as analyzing x-rays and microscopy imagery (more details about the Harbor framework are provided later in the paper). The harness preserves the standard Terminus-2 execution loop, but when a task specifies a reference file, it downloads that file from the environment and supplies it to the model in the first turn as native multimodal input. Images are attached directly, and common text-based files are inlined as text. This modification improves evaluation fidelity on tasks requiring visual or document understanding by avoiding extra turns spent on indirect file inspection, and better aligns agent evaluation with standard multimodal LLM evaluation. Evaluations run in a \href{https://modal.com/}{Modal}-provided sandbox with a controlled runtime envelope:

\begin{itemize}[nosep]
  \item \texttt{timeout\_sec = 3600.0} for the agent loop
  \item \texttt{build\_timeout\_sec = 1200.0}
  \item \texttt{cpus = 1}
  \item \texttt{memory\_mb = 2048}
  \item \texttt{storage\_mb = 10240}
\end{itemize}

\vspace{0.8\baselineskip}
The Harbor task image is bootstrapped from the following base Dockerfile:
\begin{lstlisting}
FROM python:3.12-slim

RUN apt-get update \
    && apt-get install -y --no-install-recommends \
        bash \
        tmux \
        asciinema \
        curl \
        ripgrep \
        git \
    && rm -rf /var/lib/apt/lists/*

RUN python -m pip install --no-cache-dir "litellm>=1.67.0"

WORKDIR /app

# Ensure reference assets are available inside the container.
COPY refs /app/refs

# Default shell
CMD ["/bin/bash"]
\end{lstlisting}

Verification combines exact-match checks with semantic rubric grading using an LLM jury (more details explained below in AsymmetryZero Grading Protocol). In COMPOSITE-STEM (n=70), 35 tasks are graded with exact match, 34 use semantic LLM-jury grading, and 1 uses a hybrid setup. Experts design rubrics as sets of criteria that are graded either by exact-match parsing or via an LLM-as-a-jury for semantic correctness; rubric size ranges from 1 to 40 criteria, with an average of 2.6 criteria per rubric.
The LLM jury is composed of:
\begin{itemize}[nosep]
  \item \texttt{deepseek/deepseek-v3.2}
  \item \texttt{z-ai/glm-5}
  \item \texttt{openai/gpt-oss-120b}
  \item \texttt{meta-llama/llama-3.3-70b-instruct}
  \item \texttt{moonshotai/kimi-k2.5}
\end{itemize}

At a high level, \texttt{portex\_grade.py} loads task config and criteria, reads the submission, and grades each criterion either via exact-match extraction or via multi-judge LLM scoring. It applies strict majority voting for semantic criteria, awards weighted points, aggregates to a raw score, normalizes to Harbor's reward scalar, and writes both \texttt{reward.json} and a detailed \texttt{portex\_detail.json}. The grader includes robust request handling: retries transient judge failures a limited number of times with short delays, fails fast on clear client errors, and maps persistent judge failures to explicit error outcomes with zero score so majority voting and verifier rewards stay well-defined.

The environment is a bare Python-based instance and we observe all agents using relevant packages for scientific computation and exploration such as rdkit-pypi, scipy, rd-kit, sympy, googlesearch-python, numpy, and others.

Agents are allowed to effectively use web search via installed packages like DuckDuckGo but there is no native websearch tool in the Terminus-2 harness we use. We do not supply any additional custom tooling beyond the base Dockerfile.

\end{multicols}
\begin{multicols}{2}
\section{Task Curation and Portex Datalab}

Task development was conducted in the \textbf{Portex Datalab} (\url{https://datalab.portexai.com}), where experts iterated on prompts, rubrics, and scoring assumptions while inspecting frontier-model behavior in near real time. The Portex team spoke extensively with contributors, collaborated with them to design and refine their evals, tested tasks against state-of-the-art models to gauge difficulty, and returned detailed reports and trial data so experts could iterate on task wording, rubric structure, and latent errors. This yielded a structured, multi-round curation loop instead of one-shot task drafting.

The curation workflow on the Datalab combined:
\begin{itemize}[nosep]
  \item \textbf{Eval builder workflow:} experts drafted tasks, attached references, and specified weighted criteria with explicit grading intent.
  \item \textbf{Live model feedback:} experts observed frontier-model outcomes and failure patterns while iterating on task clarity and discriminative power.
  \item \textbf{Portex review loop:} contributors received detailed run reports, artifacts, and error signals that helped them debug task specs, tighten rubrics, and correct edge-case mistakes.
  \item \textbf{Public Leaderboards:} shared leaderboards introduced light gamification that encouraged repeated quality improvements and sharper task specifications.
  \item \textbf{Cross-domain consistency:} a common Harbor execution substrate reduced harness variance while preserving domain-specific task content.
\end{itemize}
\end{multicols}

\begin{figure}[H]
\centering
\begin{minipage}[t]{0.49\textwidth}
\centering
\includegraphics[width=\linewidth]{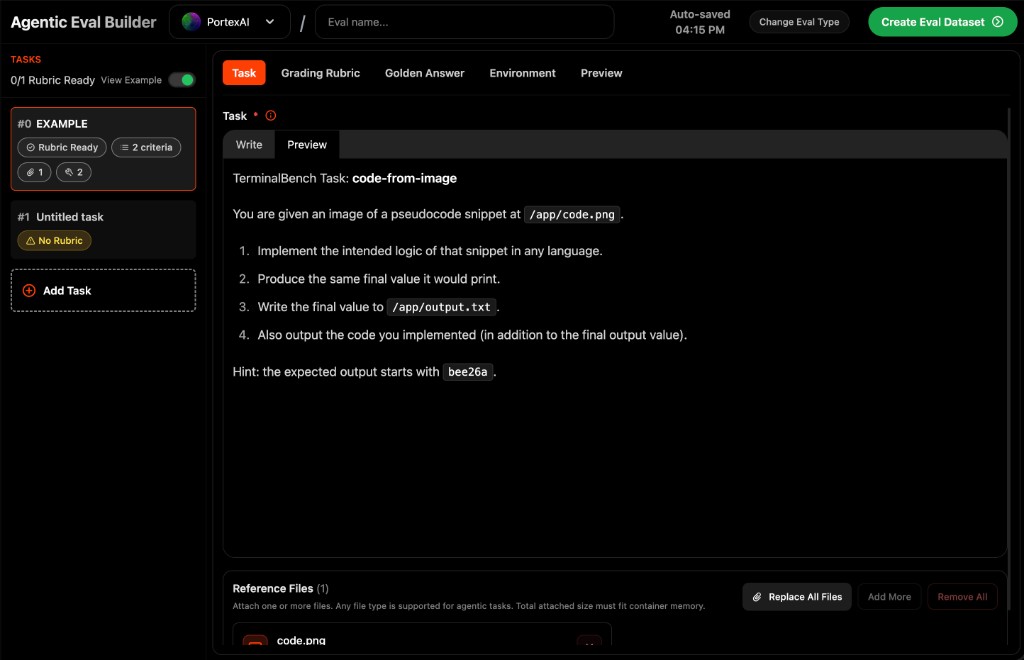}
\captionof{figure}{Portex Datalab eval-builder interface used by contributors to draft task instructions, attach reference files, and define grading rubrics.}
\end{minipage}\hfill
\begin{minipage}[t]{0.49\textwidth}
\centering
\includegraphics[width=\linewidth]{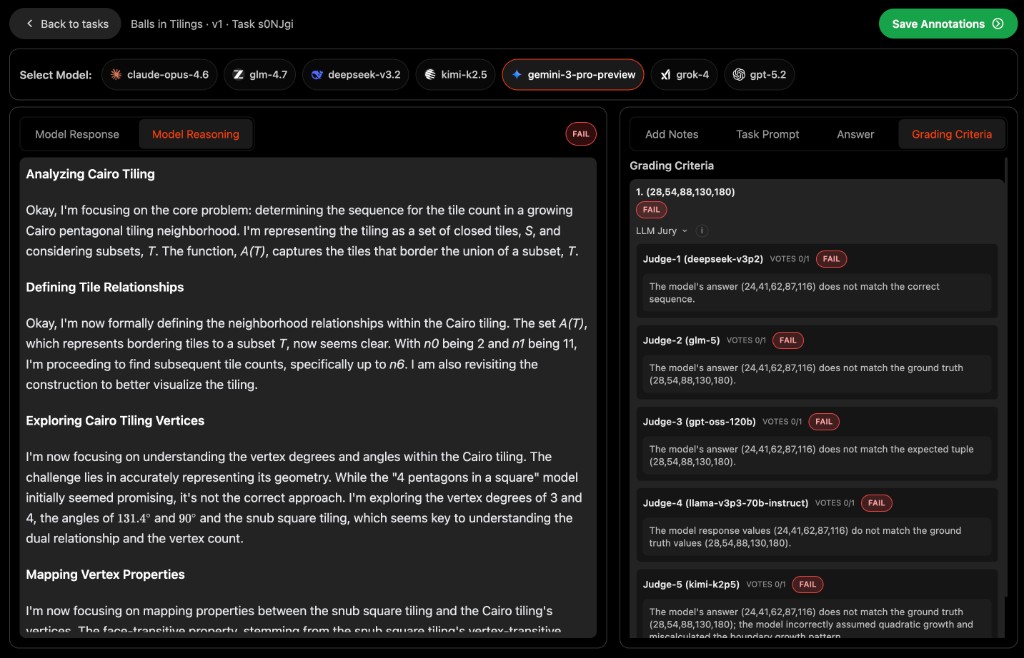}
\captionof{figure}{Portex Datalab audit view showing model traces, jury decisions, and criterion-level outcomes used to inspect failures and refine evals.}
\end{minipage}
\end{figure}

\begin{multicols}{2}

\subsection{Domain Expert Backgrounds and Task Design}

Here, we summarize expert credentials and task-design coverage for each domain.

\subsection{Physics}

Physics contributors include advanced profiles across theoretical and experimental research, including doctoral and postdoctoral work in mathematical physics, string theory, condensed matter, and quantum systems, with affiliations spanning leading universities and research institutes. Task types include symbolic derivations, quantum/open-system reasoning, many-body and topological analyses, and device-level or measurement-grounded calculations.

\subsection{Chemistry}

Chemistry contributors include researchers and practitioners in organometallic, medicinal, analytical, and computational chemistry from both academic and applied R\&D settings. Task types include synthetic-mechanism reasoning, representation conversion (e.g., SMILES/InChI), spectroscopy interpretation (NMR/MS), and concept-heavy chemical judgment.

\subsection{Biology}

Biology contributors include experts in infection biology, biomedical engineering, medical imaging, and microscopy-driven interpretation with faculty/postdoctoral and industry-adjacent backgrounds, including wet-lab settings. Task types include imaging diagnosis, MRI engineering reasoning, electron microscopy interpretation from labs, spatial-structure inference, and mechanism-aware biological analysis.

\subsection{Mathematics}

Math contributors include doctoral-level researchers and academics across pure and applied mathematics, theoretical computer science, and statistics with records in peer-reviewed venues. Task types include proof-oriented derivations, combinatorial and algebraic reasoning, stochastic-process analysis, and invariant/structure computation.

\begin{figure}[H]
\centering
\includegraphics[width=0.5\textwidth]{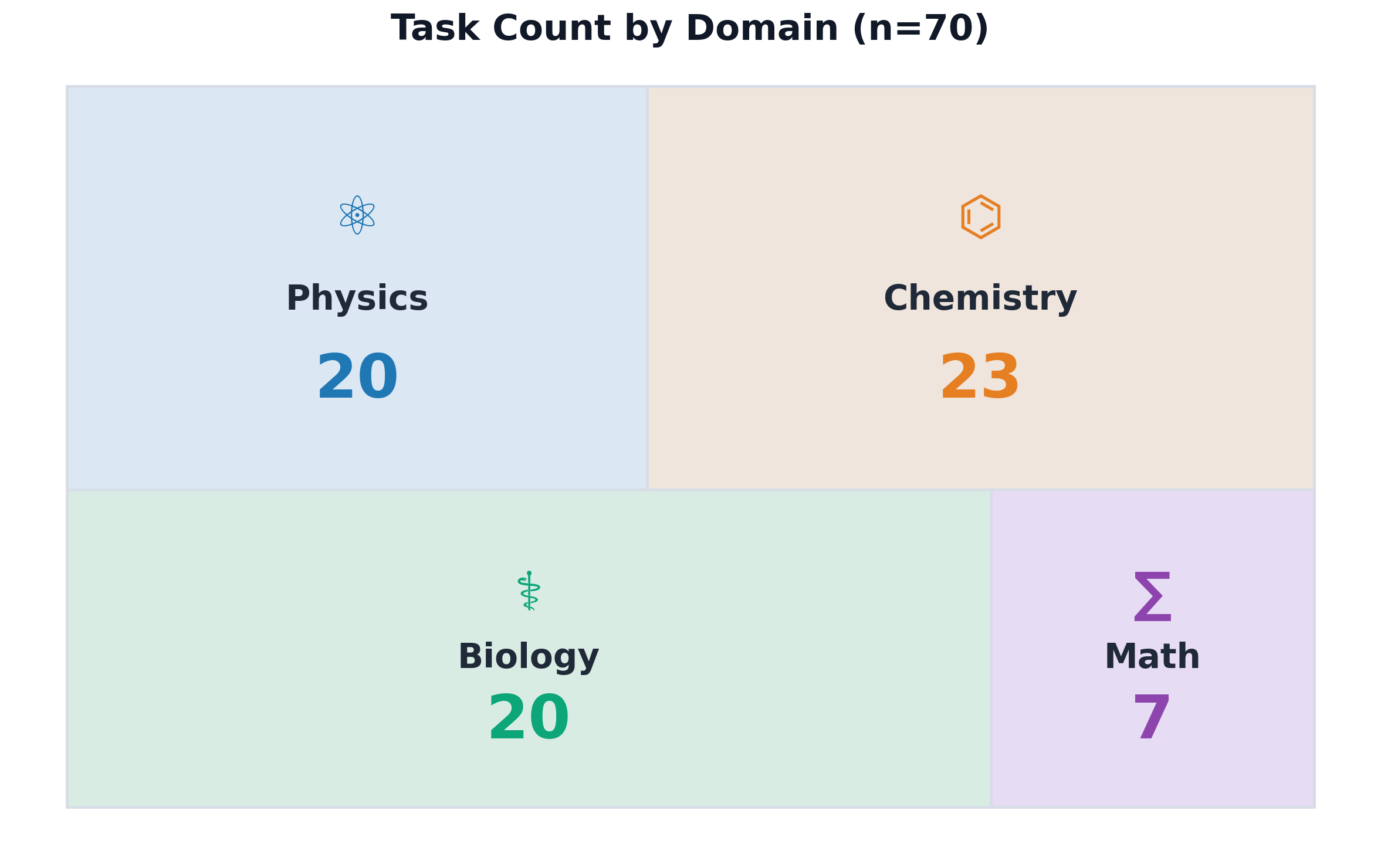}
\caption{COMPOSITE-STEM domain-level task counts.}
\end{figure}

\vspace{0.2em}

\subsection{AsymmetryZero Grading Protocol}

All grading in COMPOSITE-STEM is graded using the AsymmetryZero framework \citet{asymmetryzero}, an evaluation protocol we previously developed to operationalize expert grading preferences as stable, auditable semantic contracts. AsymmetryZero was designed to address a core challenge in benchmark design. When tasks admit multiple valid outputs or require subjective domain judgment, exact-match grading alone tends to be insufficient. At the same time, open-ended LLM judging leaves the actual grading policy implicit and difficult to audit. AsymmetryZero addresses this by representing each task as a portable evaluation contract that makes grading criteria explicit: what is being checked, how each criterion is judged, how criterion-level decisions are aggregated into a task outcome, and what threshold constitutes a pass.

Each contract is criterion-centric rather than task-centric. Every criterion declares its own weight on the task's scoring scale, a \texttt{grader\_type} (\texttt{ExactMatch} or \texttt{llm-judge}), and a criterion-specific grading instruction (\texttt{semanticPrompt}) when semantic judgment is required. This design allows deterministic and jury-based criteria to coexist inside the same task while preserving a single, well-defined aggregation rule. Criterion weights are normalized to a 0--100 scale during bundle preparation so that task scores remain comparable even when rubric sizes vary across the benchmark.

For criteria graded as \texttt{ExactMatch}, the grader extracts and normalizes the terminal answer from the agent submission and compares it against the criterion's configured reference text. For \texttt{llm-judge} criteria, each judge in the panel independently receives the task prompt, the candidate submission, and the criterion-specific semantic instruction, and returns a binary pass/fail grade together with an optional rationale. This judge aggregation policy reduces the influence of any single judge's idiosyncratic behavior on the final criterion outcome.

In COMPOSITE-STEM, the five-model frontier jury used for \texttt{llm-judge} criteria consists of DeepSeek-v3.2, GLM-5, GPT-oss-120b, Llama-3.3-70b-instruct, and Kimi-K2.5, all accessed through OpenRouter endpoints. The same AsymmetryZero evaluation contract is applied whether a task is executed via the Inspect harness for model-only runs or through Harbor for agentic evaluation, ensuring that scores remain directly comparable across execution modes. A reference implementation of the framework is open source under the MIT License at \url{https://github.com/portex-ai/asymmetry_zero}.

\subsection{Harbor Evaluation Substrate}

Harbor is a framework for agent evaluation developed by the team behind the TerminalBench benchmark. COMPOSITE-STEM is implemented on Harbor's task-and-trial abstraction so each evaluation unit is packaged and executed under a consistent agent/verifier protocol \citet{harbor}. In practice, each task includes an instruction layer, executable verification assets, and environment definitions. This lets us separate task authoring from execution orchestration while preserving a common contract for scoring and artifacts.

Task prompts are stored in \texttt{instruction.md}. Rubrics are stored as structured criteria in \texttt{tests/criteria.json}. A representative criterion specifys a criterion name, \texttt{grader\_type = llm-judge}, and a domain-specific \texttt{semanticPrompt} describing the expected derivation. We intentionally keep these rubric criteria explicit so grader behavior is auditable at the criterion level. The diagram below offers a visual guide to Harbor agent evaluation runs.

At runtime, Harbor provisions a sandbox for each trial, runs the agent interaction loop, and then runs verifier checks to produce reward files and logs. We primarily use Modal-backed execution for scalable single-container runs. This structure is important for auditability as the task spec, runtime configuration, and verifier outputs are all explicit and inspectable.

Portex grading runs through Harbor's verifier contract: the submission file (typically \texttt{/app/answer.txt}) is evaluated against all criteria, per-criterion votes are aggregated, and the verifier writes both \texttt{reward.json} (scalar reward) and \texttt{portex\_detail.json} (full breakdown with criteria results, judge votes, and explanations). This makes trial-level outcomes reproducible while preserving fine-grained diagnostics for error analysis.

For the full Harbor evaluation pipeline implementation, see the \href{https://github.com/portex-ai/portex-composit-harbor-adapter}{COMPOSITE-to-Harbor adapter repository}. It contains source-bundle retrieval, per-task dataset generation, Harbor run orchestration with configurable agents and models, optional multimodal Terminus-2 support, and the judging and verifier runtime used for grading.

\end{multicols}
\begin{center}
\resizebox{0.96\textwidth}{!}{
\begin{tikzpicture}[x=1cm,y=1cm,font=\scriptsize]
  % Header ribbon
  \node[draw, rounded corners=2pt, fill=portexlight, draw=portexgray, minimum width=12.2cm, minimum height=0.72cm, align=left] (hdr) at (6.1,5.55) {\texttt{harbor run -d <dataset> -a <agent> -m <model> [--env docker|daytona|modal]}};

  % Stage 1
  \node[draw, rounded corners=3pt, fill=blue!8, draw=blue!55!black, minimum width=3.75cm, minimum height=3.45cm, text width=3.35cm, align=left, anchor=north west, inner sep=5.5pt] (s1) at (0.1,4.95) {
    \textbf{1) Harbor task spec}\\[0.18em]
    \texttt{instruction.md}\\
    \texttt{task.toml}\\
    \texttt{app/refs}\\
    \texttt{tests/test.sh}\\
    \texttt{solution/solve.sh (opt.)}
  };

  % Stage 2
  \node[draw, rounded corners=3pt, fill=portexwarn!12, draw=portexaccent, minimum width=4.1cm, minimum height=3.45cm, text width=3.7cm, align=left, anchor=north west, inner sep=5.5pt] (s2) at (4.25,4.95) {
    \textbf{2) Sandboxed trial}\\[0.12em]
    \fbox{\begin{minipage}{3.45cm}
      \textbf{Agent run}\\
      reads \texttt{instruction.md}\\
      acts inside \texttt{/app}\\
      emits logs + trajectory
    \end{minipage}}\\[0.14em]
    \fbox{\begin{minipage}{3.45cm}
      \textbf{Verifier run}\\
      executes \texttt{/tests/test.sh}\\
      writes \texttt{reward.* + logs}
    \end{minipage}}
  };

  % Stage 3
  \node[draw, rounded corners=3pt, fill=green!11, draw=green!45!black, minimum width=3.75cm, minimum height=3.45cm, text width=3.35cm, align=left, anchor=north west, inner sep=5.5pt] (s3) at (8.75,4.95) {
    \textbf{3) Outputs + analysis}\\[0.18em]
    \texttt{result.json}\\
    \texttt{agent/trajectory.json}\\
    \texttt{verifier/reward.* + logs}\\
    \texttt{artifacts/}\\[0.22em]
    \fbox{\begin{minipage}{3.05cm}
      \centering \textbf{Harbor viewer}\\
      trials, trajectories, rewards
    \end{minipage}}
  };

  % Arrows
  \draw[->, line width=0.9pt, color=portexaccent] (s1.east) -- (s2.west);
  \draw[->, line width=0.9pt, color=portexaccent] (s2.east) -- (s3.west);

  % Optional module note
  \node[draw, dashed, rounded corners=2pt, fill=white, draw=portexgray, minimum width=4.8cm, minimum height=0.68cm, align=center] at (6.15,0.62) {\textit{Optional: docker-compose sidecars / extra artifact paths}};
\end{tikzpicture}
}\\[-0.15em]
\footnotesize\textit{Harbor turns task specs into reproducible sandboxed trials, verifier-scored rewards, and inspectable outputs.}
\end{center}
\section{Model Results}

\subsection{Compact Leaderboard}
\begin{center}
\scriptsize

\begin{table}[H]
\centering
\scriptsize
\caption{Performance metrics by model.}
\begin{tabular}{l c c c c c c}
\toprule
\textbf{Model} & \textbf{Harness} & \textbf{Pass@1} & \textbf{Avg Time} & \textbf{Avg Episodes} & \textbf{In Tok} & \textbf{Out Tok} \\
\midrule
\includegraphics[height=0.25cm]{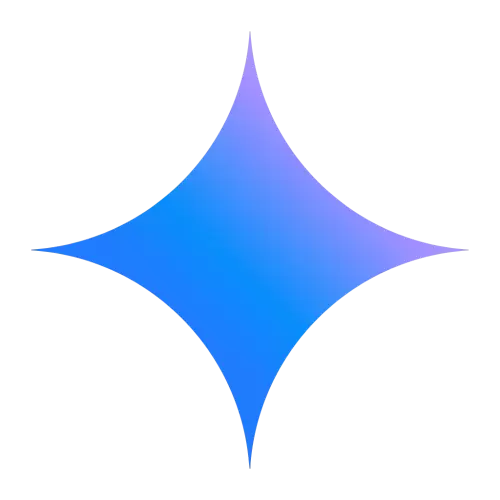}~\texttt{gemini-3.1-pro} & \texttt{Terminus-2} & 18.6\% & 9m 16s & 6.7 & 36.2K & 18.4K \\
\includegraphics[height=0.25cm]{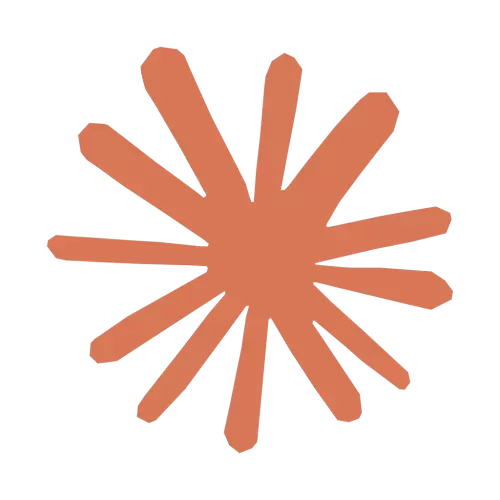}~\texttt{claude-opus-4.6} & \texttt{Terminus-2} & 21.4\% & 11m 30s & 5.6 & 156.0K & 21.1K \\
\includegraphics[height=0.25cm]{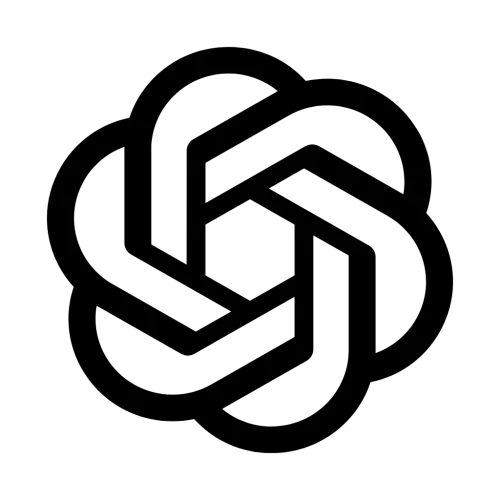}~\texttt{gpt-5.4} & \texttt{Terminus-2} & 4.3\% & 6m 5s & 2.6 & 17.4K & 2.4K \\
\includegraphics[height=0.25cm]{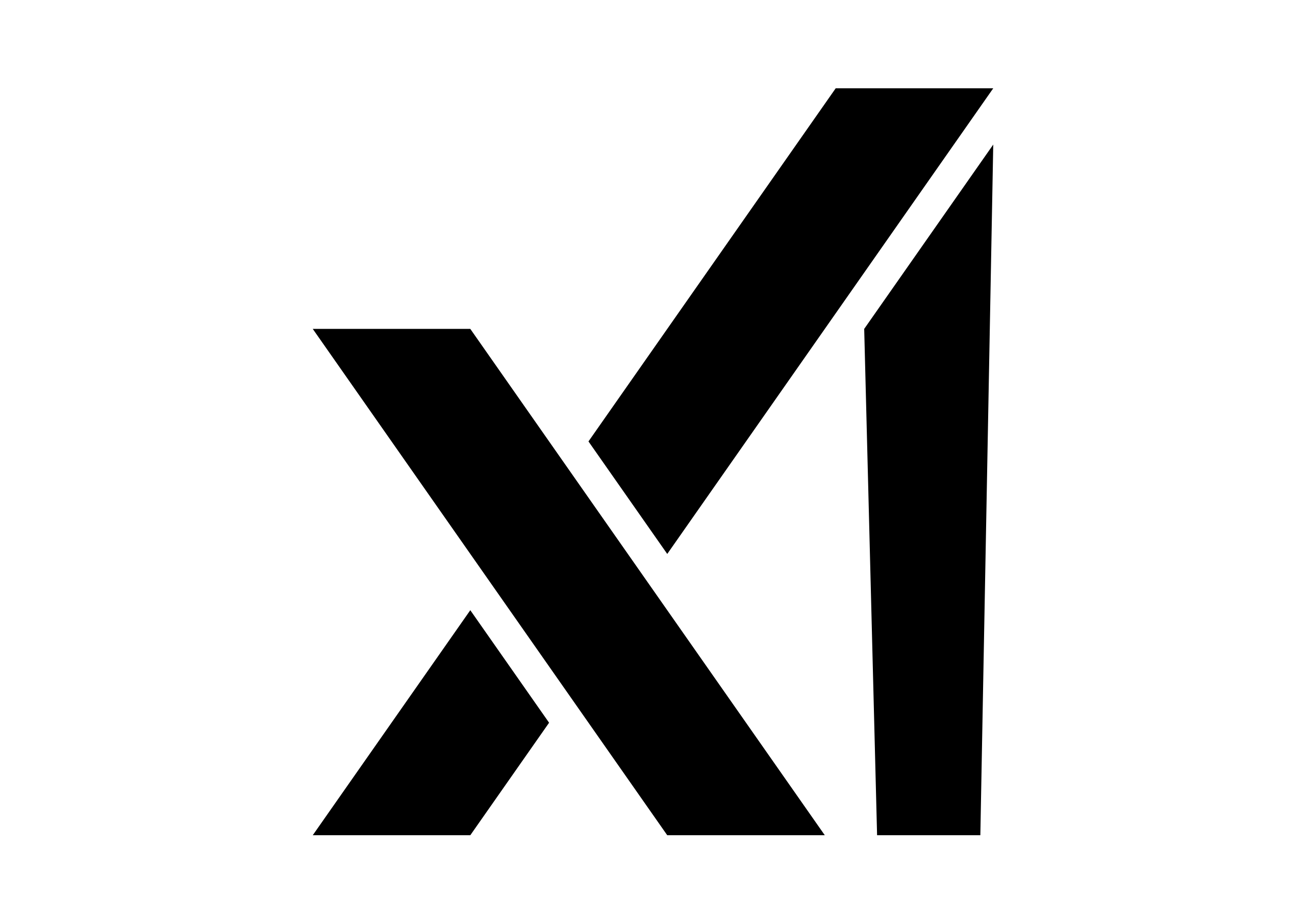}~\texttt{grok-4.20-beta} & \texttt{Terminus-2} & 5.7\% & 6m 3s & 3.3 & 20.6K & 4.7K \\
\bottomrule
\end{tabular}
\end{table}

\end{center}
\subsection{Task-by-Model Pass Matrix}
\begin{figure}[H]
\centering
\includegraphics[width=0.99\textwidth]{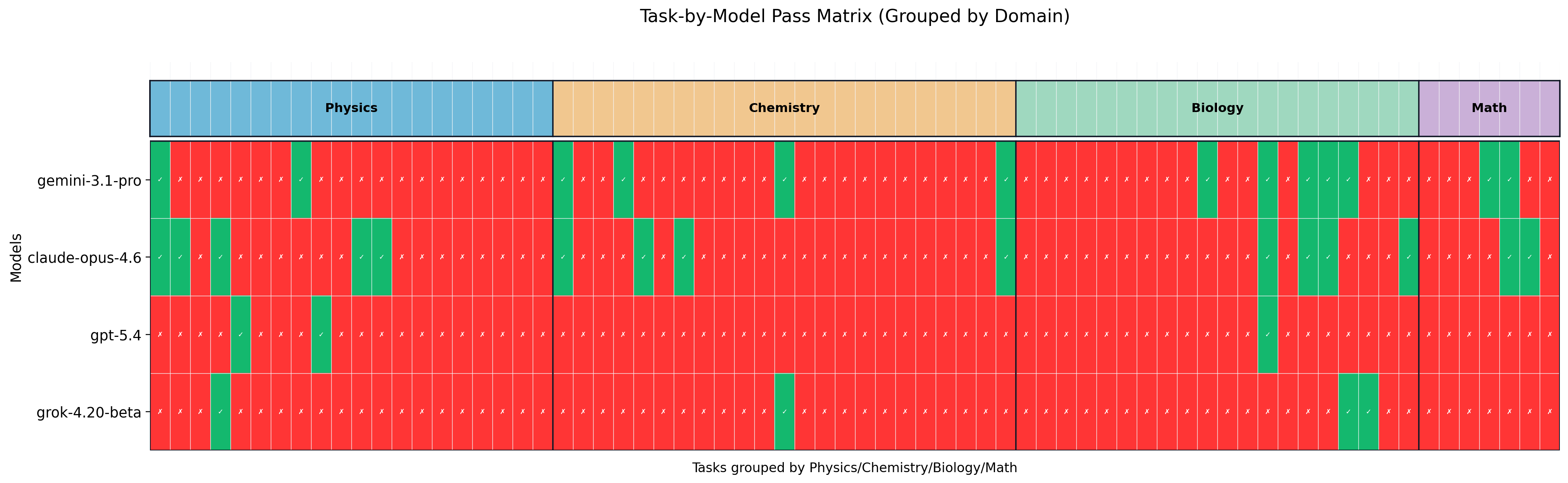}
\caption{Dense task-model outcome heatmap grouped by domain (green=pass, red=fail, gray=not scored).}
\end{figure}

\begin{multicols}{2}
\section{Pass@1 Failure Taxonomy}
To better characterize \emph{why} models fail on COMPOSITE-STEM, we report a compact two-way taxonomy over failed trials using raw Harbor artifacts such as \texttt{result.json}, \texttt{reward.json}, \texttt{portex\_detail.json}, and \texttt{trajectory.json}. \textbf{Solution Error} means an answer was submitted but graded incorrect. \textbf{Submission Error} means the run failed to deliver a valid submission artifact (missing/invalid output), and we also count any run that reached \texttt{max\_turns=10} as Submission Error by definition.

\end{multicols}

\begin{table}[t]
\centering
\footnotesize
\caption{Breakdown of Failed Trials by Type.}
\resizebox{0.98\textwidth}{!}{%
\begin{tabular}{l c c c}
\toprule
\textbf{Model} & \textbf{Failed trials} & \textbf{Solution Error} & \textbf{Submission Error} \\
\midrule
\texttt{claude-opus-4.6} & 54 & 63.0\% (34) & 37.0\% (20) \\
\texttt{gemini-3.1-pro} & 56 & 46.4\% (26) & 53.6\% (30) \\
\texttt{gpt-5.4} & 66 & 90.9\% (60) & 9.1\% (6) \\
\texttt{grok-4.20-beta} & 66 & 83.3\% (55) & 16.7\% (11) \\
\bottomrule
\end{tabular}%
}
\end{table}

\begin{multicols}{2}
Across models, \textbf{Solution Error} remains dominant for Claude, GPT, and Grok, while Gemini shows a larger \textbf{Submission Error} share. This two-category view cleanly separates incorrect solved outputs from delivery or format failures.

\paragraph{Case study: \texttt{COMPOSITE-STEM Chemistry Task ID 0sL35u} (See Appendix for full details)}
Here, we highlight a single chemistry task to demonstrate failure modes between two models, Claude Opus 4.6 and GPT 5.4. Both models faced the same initial constraint (chemistry tooling absent by default), but diverged in recovery strategy. Claude persisted on environment setup, installed \texttt{python3-rdkit} via \texttt{apt}, and used RDKit to parse the full SMILES, producing the correct answer (\texttt{350}) and passing exact-match grading. GPT attempted multiple tooling paths (\texttt{rdkit}, \texttt{openbabel}, \texttt{pip}) but then switched to a custom SMILES/valence parser; that fallback produced \texttt{Answer: 399} while the verifier reference was \texttt{350}, yielding a full fail under exact-match scoring. In plain terms, persistence toward a robust domain tool succeeded, while a fragile hand-rolled workaround failed on a high-complexity molecule. We explore broad differences in model performance below.

\section{Exploring Model Performance Differences}

One of the clearest observations in the model results is the spread between frontier models: \texttt{claude-opus-4.6} scored 21.4\% and \texttt{gemini-3.1-pro} scored 18.6\%, while \texttt{gpt-5.4} and \texttt{grok-4.20-beta} scored 4.3\% and 5.7\%. Because that gap is large, we first asked whether it could be explained by stack-level failures. On the verifier side, errors were rare: just 4 instances across all four models, a little over 1\% of runs, and not disproportionately concentrated in any one model. On the solver side, ``missing submissions'' appeared in two forms: runs that exceeded the \texttt{max\_turns=10} limit and therefore received an automatic zero, and runs that stayed within budget but still failed to produce a gradeable \texttt{answer.txt} submission. These are better understood as agent failure modes than infrastructure bugs.

By model, step-limit failures were most common for Gemini (26 runs), followed by Opus (6), GPT (3), and Grok (1); malformed or ungradeable submissions were less common, with 1 for Opus, 3 for GPT, and 2 for Grok. So while solver-side omissions were present, they were not large enough to explain the ranking and, if anything, Gemini was the model most affected by them. More importantly, the broader pattern still held after controlling for these omissions and errors. When we filtered out non-gradeable runs, the gap not only remained but widened: the grouped summaries show an overall score gap of 0.191, but a non-error score gap of 0.280, with the stronger models averaging roughly 0.37 on non-error runs versus roughly 0.09 for GPT/Grok. That pattern suggests the observed difference is unlikely to be primarily an infrastructure artifact; if anything, the underlying performance gap becomes more visible once stack failures are removed.

Because much of the benchmark relies on LLM-judge criteria, we also looked for signs that grader behavior might favor some models over others. The evidence points in the opposite direction. The Anthropic/Google models (strong group) collectively show more 3--2 splits and higher entropy than the Grok/GPT models, which mostly register unanimous failures (0--5). The weak group's combined stats are 166/190 unanimous, 183/190 near-unanimous, and 6 non-trivial splits, whereas the strong group reaches only 127/195 unanimous, with 18 splits whose entropy exceeds 0.97. This suggests the stronger models generate answers that sit closer to the pass/fail margin, leading to higher judge disagreement. Lower-performing models are judged more consistently because they tend to produce outright incorrect answers that all five judges reject. The higher disagreement for Opus/Gemini reflects partially correct or borderline responses, not a grading bias that artificially suppresses GPT/Grok.

\end{multicols}
\begin{table}[H]
\centering
\footnotesize
\resizebox{0.98\textwidth}{!}{%
\begin{tabular}{l c c c c c c c}
\toprule
\textbf{Model} & \textbf{Combos} & \textbf{Unanimous} & \textbf{Near-unanimous} & \textbf{Avg. entropy} & \textbf{5--0 / 0--5} & \textbf{4--1 / 1--4} & \textbf{3--2 / 2--3} \\
\midrule
\texttt{claude-opus-4.6} & 110 & 79/110 & 101/110 & 0.224 & 30 / 49 & 6 / 16 & 6 / 9 \\
\texttt{gemini-3.1-pro} & 85 & 48/85 & 76/85 & 0.341 & 13 / 35 & 14 / 14 & 3 / 6 \\
\texttt{gpt-5.4} & 93 & 79/93 & 87/93 & 0.125 & 11 / 68 & 1 / 7 & 2 / 4 \\
\texttt{grok-4.20-beta} & 97 & 87/97 & 96/97 & 0.077 & 8 / 79 & 3 / 6 & 1 / 1 \\
\bottomrule
\end{tabular}
}
\caption{Judge-vote agreement patterns by model across LLM-judge criteria combinations.}
\end{table}

\begin{multicols}{2}
That leaves the question of what might explain the remaining difference. One suggestive pattern is solver behavior. The stronger models appear to use more of the available step budget and engage tools more often: Opus averages about 6.5 total steps and Gemini about 8, compared with roughly 3.8 for GPT and 4.5 for Grok. Tool-use traces point in the same direction. Opus and Gemini install packages in 37 runs combined, about 30\% of their runs, while GPT and Grok do so only 7 times total, and GPT/Grok overwhelmingly fall into the ``no substantive tool use'' category (66 and 63 runs, respectively). Those minimal-tool runs also have very low pass rates, around 3--5\%, and tend to end after only 3--4 steps, often with a direct write to \texttt{/app/answer.txt}. By contrast, runs involving more exploration or tool use average closer to 10 steps and show modestly better success. These are only observational signals, but they suggest that part of the gap may come from how models allocate effort: Opus and Gemini appear more likely to keep the loop alive, inspect intermediate outputs, and use tools, and that behavior correlates with better outcomes on these tasks.

\end{multicols}
\begin{table}[H]
\centering
\footnotesize
\begin{tabularx}{\textwidth}{@{}l l c c c@{}}
\toprule
\textbf{Model} & \textbf{Label} & \textbf{Runs} & \textbf{Pass rate} & \textbf{Avg. total / agent steps} \\
\midrule
\texttt{claude-opus-4.6} & installs\_package & 13 & 23\% & 10.9 / 9.9 \\
\texttt{claude-opus-4.6} & no\_substantive\_tool\_use & 57 & 21\% & 5.3 / 4.3 \\
\texttt{gemini-3.1-pro} & installs\_package & 24 & 17\% & 10.5 / 9.5 \\
\texttt{gemini-3.1-pro} & calls\_preinstalled\_tools & 2 & 0\% & 9.5 / 8.5 \\
\texttt{gemini-3.1-pro} & no\_substantive\_tool\_use & 44 & 20\% & 6.1 / 5.1 \\
\texttt{gpt-5.4} & installs\_package & 1 & 0\% & 6.0 / 5.0 \\
\texttt{gpt-5.4} & calls\_preinstalled\_tools & 3 & 0\% & 11.0 / 10.0 \\
\texttt{gpt-5.4} & no\_substantive\_tool\_use & 66 & 5\% & 3.3 / 2.3 \\
\texttt{grok-4.20-beta} & installs\_package & 6 & 17\% & 7.3 / 6.3 \\
\texttt{grok-4.20-beta} & calls\_preinstalled\_tools & 1 & 0\% & 3.0 / 2.0 \\
\texttt{grok-4.20-beta} & no\_substantive\_tool\_use & 63 & 3\% & 4.2 / 3.2 \\
\bottomrule
\end{tabularx}
\caption{Observed solver-behavior clusters and associated Pass@1 outcomes.}
\end{table}

\begin{multicols}{2}
\section{Limitations}

This benchmark was built with strong expert involvement, but it did not undergo a full external audit and quality-control and peer-review pipeline equivalent to a formal academic benchmark consortium. While this is a limitation, the curation process included multiple live calls with experts, repeated back-and-forth on eval design, and iterative task revisions after observing model behavior. We also stress that almost all contributors underwent and passed extensive auditing and vetting with previous benchmarking efforts, such as Humanity's Last Exam (HLE).

We did not observe evidence of misaligned contributor incentives. In practice, much of the motivation appeared to be scientific curiosity about how well frontier AI systems understand specialized domains. Experts largely approached contribution as an experiment in measuring model capability rather than as an optimization for benchmark gaming.

For this evaluation setup, agent runs were configured with a fixed interaction budget of \texttt{max\_turns=10}. If a run did not produce the required output artifact (for example, no submission written to \texttt{/app/answer.txt}) within that budget, we counted the trial as a task failure. Longer agent runs might produce different results, and we invite others to experiment with the full suite of tasks on \href{https://huggingface.co/datasets/portex/COMPOSITE-STEM/tree/main}{Hugging Face}.

We also report only Pass@1 results in this report. Because agent trajectories and judge-based grading can exhibit some natural run-to-run variation, a more statistically robust characterization would include repeated runs per model to better calibrate variance and confidence in the reported rankings.

\section{Conclusion}

Agents may markedly improve researcher efficiency in professional and scientific settings; yet, this adoption hinges upon experts confidently knowing when to entrust agents with real work. Scientific benchmarks written by experts play a key role in establishing this trust by showing where agents do well, or poorly, on real-world tasks. 

To this end, we constructed COMPOSITE-STEM, a benchmark designed to test agents in terminal environments on difficult STEM tasks across physics, biology, chemistry, and mathematics. COMPOSITE-STEM curates 70 task contributions from 20 faculty members, applied scientists, and doctoral-level researchers from leading global universities and research institutions. COMPOSITE-STEM is integrated with the Harbor framework for interoperability with leading agent benchmarks and reinforcement learning frameworks. Our use of LLM-as-a-Jury grading provides flexibility to experts to design rubrics with semantic criterion that go beyond simple exact-match grading. Our Pass@1 performance calibration shows that state-of-the-art agents have room for improvement. By open sourcing the full task dataset and Harbor adapter, we hope this benchmark can be used as another robust and independent baseline to accompany the advancement of AI agents in the sciences.

\end{multicols}
\newpage
\addcontentsline{toc}{section}{References}
\bibliography{references}

\newpage
\appendix
\section{Appendix: Case Study Trajectories}

This appendix includes details for Chemistry task ID \texttt{0sL35u}, with trajectory outcomes for \texttt{claude-opus-4.6} and \texttt{gpt-5.4}.

\subsection*{\texttt{0sL35u}: Chemistry}
\textbf{Grading mode:} Exact Match
\textbf{Full task instruction (\texttt{instruction.md}):}
\begin{lstlisting}
You are a graduate student working in the laboratory of a formerly renowned marine natural product chemist at a well-known land-grant university. You and your colleagues have been tasked with cataloguing a historical stock of secondary metabolites that have isolated over the PI's career and entering these into a modern chemical inventory system. The compounds have been entered into the system as small molecule input line entry system (SMILES) codes. One of these, reportedly isolated from a marine sponge, catches your attention with the following SMILES code: 

O[C@@H](C[C@@H](O)[C@H](OO[C@@H](CC(N[C@@H](CC(C)C)C(N[C@@H](C(N[C@H]([C@H]1C)C(N[C@@](C(C)C)([H])C(N[C@@H](CC(C)C)C(N[C@H](CO)C(N[C@@H](CC(C)C)C(N[C@H](CO)C(N[C@](CC(C)C)([H])C(O1)=O)=O)=O)=O)=O)=O)=O)=O)CCC(O)=O)=O)=O)CCCCCCCCC)CC[C@@H](O)C[C@@H](O)CC2=O)C[C@@]3(O)O[C@H](CC(O[C@@]4([H])[C@@H](OC)[C@@H](N)[C@H](OC5=CC6=C(C=C5)C(C(C7=CN(N(C8=C9C=CC(Br)=C8)C=C9C(C%10=CN(S(=O)(O)=O)C%11=C%10C=C(C%12=CC%13=C(C=C%12Br)NC=C%13C(C%14=CN(S(=O)(O)=O)C%15=C%14C=CC(Br)=C%15)=O)C(Br)=C%11)=O)C%16=C7C=CC(Br)=C%16)=O)=CN6S(=O)(O)=O)[C@@H](C)O4)/C=C/C=C/C=C/C=C/C=C(Cl)/C=C(Cl)/C=C/[C@@H](C)[C@@H](OC([C@H]%17[C@@H](CC(O[C@@]%18([H])[C@@H](OC)[C@@H](N)[C@H](O)[C@@H](C)O%18)/C=C/C=C(C%19=C%20C([C@@]%21%22C(N%20)=C(C)C[C@@]%23(CC)[C@@H]%21N(CC%22)C[C@H]%24[C@@H]%23OC%25=C%24C=C([C@@]%26%27C(N%28)=C(C(OC)=O)C[C@@]%29(CC)[C@@H]%26N(CC%27)CC=C%29)C%28=C%25)=CC([C@@H]%30[C@@H]%31[C@@H](O%31)[C@@]%32(CC)CC(C(OC)=O)=C%33[C@]%34(C(C=C(O)C(OC)=C%35OC)=C%35N%33)[C@H]%32N%30CC%34)=C%19OC)/C=C/C=C/C=C(I)/C=C(I)/C=C/[C@@H](C)[C@@H](O)[C@@H](O)[C@H](O)OC%36=O)O[C@@](C[C@@H](O)C[C@@H](O)[C@H]%37CC([C@@H](O)C[C@@H](O)C%36)=C(CCCCCCCCC%38=CC=CC=C%38)O%37)(O)C[C@@H]%17O)=O)[C@@H](O)[C@H](O)O2)[C@H](C(OC)=O)[C@@H](O)C3

You happen to have a passing interest in cheminformatics, as well as logical deduction, so you attempt to decipher the structure of the metabolite from the SMILES code. For the purpose of this exercise, determine how many hydrogen atoms there are on the molecule.


Reference file path: `(none)`

After your reasoning, state your final answer clearly on a single line that starts with Answer: so it can be parsed automatically. Example: `Answer: The capital of France is Paris.`

Write your complete response (including reasoning and final answer) to `/app/answer.txt`.
\end{lstlisting}

\textbf{Submitted answer (\texttt{claude-opus-4.6}):}
\begin{lstlisting}
Using RDKit to parse the SMILES string, the molecule has the molecular formula C280H350Br5Cl2I2N23O75S3.

After adding explicit hydrogens to the parsed molecule, counting all atoms with atomic number 1 gives us the total number of hydrogen atoms.

The molecule has 740 total atoms (390 heavy atoms + 350 hydrogen atoms).

Answer: 350
\end{lstlisting}

\textbf{Submitted answer (\texttt{gpt-5.4}):}
\begin{lstlisting}
I determined the hydrogen count by parsing the provided SMILES with a custom Python script. The script tokenized the SMILES, built the atom/bond connectivity including branches and ring closures, summed bond orders for each atom, and then assigned hydrogens using standard valences for the elements present (C 4, N 3, O 2, halogens 1, sulfur as needed for sulfone valence). It also counted the four explicitly written [H] atoms already present in the SMILES.

The script reported these atom counts:
- C: 280
- H (explicit in brackets): 4
- N: 23
- O: 75
- S: 3
- Br: 5
- Cl: 2
- I: 2

It then computed the total number of hydrogen atoms as 399, with no valence inconsistencies detected.

Answer: 399
\end{lstlisting}

\begin{table}[H]
\centering
\small
\begin{tabularx}{\textwidth}{@{}l c c c X@{}}
\toprule
\textbf{Model} & \textbf{Score} & \textbf{Result} & \textbf{Episodes} & \textbf{Trajectory summary} \\
\midrule
\texttt{claude-opus-4.6} & 100\% & \textcolor{portexsuccess}{PASS} & 10 & Attempted Python package routes, then successfully installed \texttt{python3-rdkit} via apt; used RDKit-based parsing to compute molecular formula and hydrogen count; wrote exact final answer (\texttt{350}) to \texttt{/app/answer.txt}. \\
\texttt{gpt-5.4} & 0\% & \textcolor{portexfail}{FAIL} & 5 & Tooling checks found no preinstalled chemistry libraries; install attempts via Python package paths failed; pivoted to a custom SMILES/valence parser and produced \texttt{399}, which failed exact-match grading against the reference answer \texttt{350}. \\
\bottomrule
\end{tabularx}
\caption{Case-study trajectory comparison for task \texttt{0sL35u}.}
\end{table}

\end{document}